\documentclass{cta-author}

\usepackage{algorithm}
\usepackage{algorithmic}
\usepackage{graphics,epsfig, subfigure}
\usepackage{caption}
\usepackage[most]{tcolorbox}
\usepackage{float}

\usepackage{amsthm}
\usepackage{amsmath,amssymb}
\usepackage{amsfonts}
\usepackage{array}
\usepackage{tabu}
\usepackage{enumitem}
\usepackage{epstopdf}
\usepackage{appendix}
\graphicspath{ {figs/} }
\usepackage{multirow}

{}
{}
{}

\begin{document}


\title{Three Dimensional Route Planning for Multiple Unmanned Aerial Vehicles using Salp Swarm Algorithm}


\author{\au{Priyansh Saxena$^{\corr}$}, \au{Ram Kishan Dewangan$^{\corr}$}}

\address{{$^{\corr}$ABV - Indian Institute of Information Technology and Management Gwalior, MP, India
}\\
\email{saxenapriyanshasd@gmail.com}}

\begin{abstract}
Route planning for multiple Unmanned Aerial Vehicles (UAVs) is a series of translation and rotational steps from a given start location to the destination goal location. The goal of the route planning problem is to determine the most optimal route avoiding any collisions with the obstacles present in the environment. Route planning is an NP-hard optimization problem. 
In this paper, a newly proposed Salp Swarm Algorithm (SSA) is used, and its performance is compared with deterministic and other Nature-Inspired Algorithms (NIAs). 
The results illustrate that SSA outperforms all the other meta-heuristic algorithms in route planning for multiple UAVs in a 3D environment. The proposed approach improves the average cost and overall time by $1.25\%$ and $6.035\%$ respectively when compared to recently reported data. 
Route planning is involved in many real-life applications like robot navigation, self-driving car, autonomous UAV for search and rescue operations in dangerous ground-zero situations, civilian surveillance, military combat and even commercial services like package delivery by drones.\\


\end{abstract}

\maketitle

\section{Introduction}\label{sec1}

UAVs have the capability of autonomous navigation which allows them to move towards the goal location in the most optimal fashion, and simultaneously ensure that they do not suffer any collisions with other UAVs or the obstacles present in the environment. The most compelling ability of the UAVs is to operate in complex environments where human operations tend to be very difficult. The time needed to solve the route planning problem is exponential and substantially grows with the increasing complexity of environment since it is an NP-hard problem.

There are two different methods for route planning \cite{ref-a}: \textit{offline (or global) route planning} -- in which the UAV estimates the complete route even before starting any movement; and \textit{online (or local) route planning} -- in which the vehicle simultaneously updates the route and moves towards the destination.

Route planning can be applied to land, air or water -- while the vehicle in land corresponds to a 2D environment; in the latter two, the UAV can move in all the three dimensions and corresponds to a 3D environment. This paper considers the problem of global route planning for multiple UAVs in 3D environment.

\begin{figure}[h]
\centering
\includegraphics[width=85mm,scale=0.85]{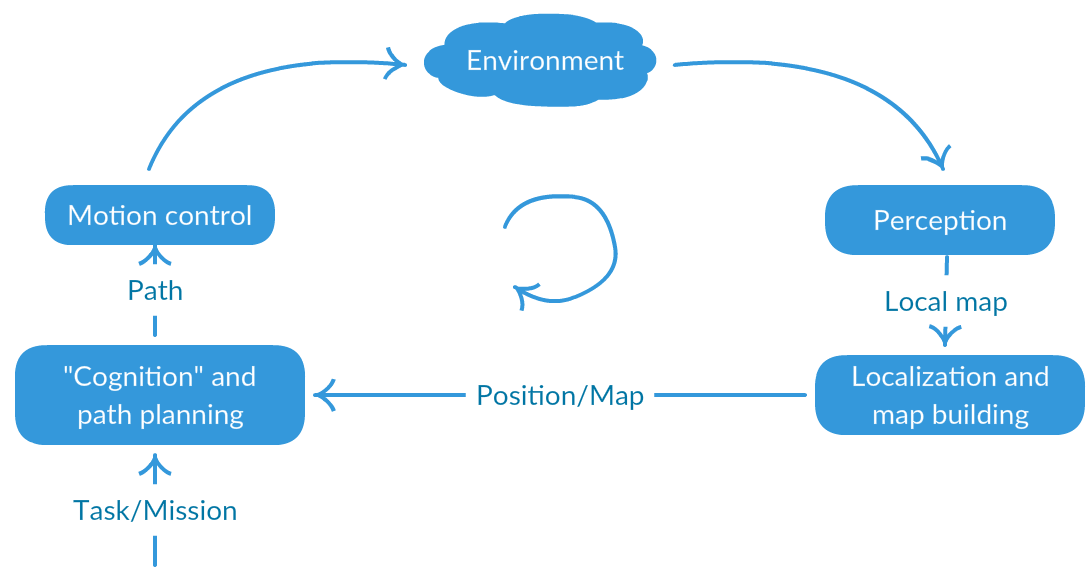}
\caption{Robot Navigation Structure \cite{ref-w}}
\label{fig:1}
\end{figure}

As shown in Figure \ref{fig:1}, the underlying architecture of route planning can be divided into four components: 

\begin{enumerate}
\item \textbf{Perception}: Vehicle utilizes the sensors to devise meaningful information of the surroundings. If the agent/robot possesses full knowledge of the environment at all time then the route planning is global otherwise it is local.
\item \textbf{Localization}:  Vehicle identifies its location in the operating environment. 
\item \textbf{Cognition and path planning}: Vehicles decide in which direction it should steer to reach to the goal location in accordance with the deterministic or meta-heuristic algorithm used.
\item \textbf{Motion control}: Vehicle regulates its motion in order to achieve the desired trajectory. 
\end{enumerate}

Meta-heuristic algorithms can be termed as stochastic algorithms with randomization and local search \cite{ref-a}. The main reason to choose a meta-heuristic based SSA algorithm for route planning is that these algorithms generate near-optimal routes in significantly less time in complex environments which can not be achieved by deterministic algorithms. Meta-heuristic algorithms are very efficient and have a wide range of applications since they can achieve practical solutions for various problems.

Meta-heuristic algorithms are classified as \textit{Evolutionary} and \textit{Swarm Intelligence}. The former algorithm tries to mimic the approach of evolution in nature; and the latter tries to mimic the intelligence of herds, swarms, flocks in nature. The primary source of inspiration for these techniques emerge from the collective behavior of creatures. Some of the evolutionary algorithms include Genetic Algorithm, Differential evolution, Biogeography-Based Optimization (BBO), Evolution strategy; while swarm intelligence includes Ant Colony Optimization (ACO), Particle Swarm Optimization (PSO), Dragonfly Algorithm (DA), Cuckoo Search (CS), Grey Wolf Optimizer (GWO), Whale Optimization Algorithm (WOA), Salp Swarm Algorithm (SSA) among others.

\section{Related Work}\label{sec2}

The problem of route planning by autonomous UAVs in a 2D environment has been solved by many approaches like cell decomposition method \cite{ref-b}, Voronoi diagram, visibility graph \cite{ref-c}, potential field approach, and rapidly exploring random trees (RRTs) \cite{ref-d}, deterministic search algorithm Dijkstra \cite{ref-e} and heuristic based algorithms (A* and D*) \cite{ref-f}. 
The algorithms mentioned above are proactive, so they are not effective solutions for route planning and suffer from local minima stagnation and considerable time complexity.

In a 3D environment various meta-heuristics algorithms like Genetic Algorithm (GA) \cite{ref-g}, Predator-Prey Pigeon Inspired Optimization (PPPIO) \cite{ref-h}, Whale Optimization Algorithm (WOA) \cite{ref-i}, Biogeography-Based Optimization (BBO) \cite{ref-j}, Particle Swarm Optimization (PSO) \cite{ref-k}, Invasive Weed Optimization (IWO) \cite{ref-l}, Glowworm Swarm Optimization (GSO) \cite{ref-m} etc. have been applied. Md. Arafat \cite{ref-n} presented Bacterial Foraging Optimization (BFO) to compute the shortest path in a dynamic unknown environment and used Gaussian cost function. Similarly, Edin Dolicanin \cite{ref-o}, applied a modification of BrainStorm Optimization (BSO) Algorithm for finding the optimal path of an unmanned combat aerial vehicle (UCAV) while considering fuel consumption and safety degree as a metric. Zhang et al. \cite{ref-p} applied Grey Wolf Optimizer (GWO) to path planning issue on the battlefield. Phung et al. \cite{ref-q} improved discrete PSO technique. He devised it to path planning for surface inspection using UAV vision. Yaoming Zhou \cite{ref-r} proposed a bio-inspired computing algorithm that is inspired from plant growth mechanism and applied it to the problem of path planning. 

In general, nature-inspired algorithms have displayed excellent performance in solving intricate real-world problems like route planning. In this paper, the extension of some above-listed algorithms for finding routes for multiple UAVs in a 3D environment is presented. This paper proposes the use of Salp Swarm Algorithm(SSA) which is a meta-heuristic optimization algorithm and proposed by Mirjalili et al. \cite{ref-s}  to solve the route planning problem. SSA is  influenced by swarming behavior of salps while foraging and navigating in oceans. In \cite{ref-s}, it was shown that SSA significantly outperforms other popular meta-heuristic algorithms. By using several stochastic operators, the problem of local optima stagnation is avoided in multi-modal search landscapes. The algorithm is applicable for single as well as multi-objective optimization tasks.

The further structure of the paper is formulated into following sections. Section \ref{sec:3} illustrates the methodology of multiple autonomous UAV route planning and a description of the environment. Section \ref{sec:4} gives a thorough explanation of the SSA. Section \ref{sec:5} illustrates the implementation and final results obtained by comparing SSA with other meta-heuristic algorithms. Section \ref{sec:6} finally concludes the paper.

\section{Methodology}\label{sec:3}

The proposed methodology can be organized into the following segments. First, a description of problem statement is given and after that terrain construction is described. Following that, the cost function for the optimization is presented, and then finally the trajectory for autonomous UAVs is generated.


\subsection{Problem Statement}

The problem of route planning of multiple UAVs in 3D environment can be viewed as a mapping to a function in which start and goal location are presented as inputs and obtain an optimal trajectory in terms of output. The primary interest of route planning is to generate an optimal, collision-free route with the least cost. 

\begin{equation}
f(start_i, goal_i) \rightarrow trajectory_i  
\label{eq:1}
\end{equation}

where $start_i$ represents start location of the $i^{th}$ UAV $(X_{start}, Y_{start}, Z_{start})$; $goal_i$ represents the goal location of the $i^{th}$ UAV $(X_{goal}, Y_{goal}, Z_{goal})$; $trajectory_i$ represents a collision-free trajectory of $i^{th}$ UAV \cite{ref-x}.

The initial position of $n$ UAVs can be represented with the help of the following matrix:

\begin{equation}
pos = \left[ {\begin{array}{*{20}{c}}
  {pos_1^1}&{pos_2^1}&{...}&{...}&{pos_n^1} \\ 
  {pos_1^2}&{pos_2^2}&{...}&{...}&{pos_n^2} \\ 
  {...}&{...}&{...}&{...}&{...} \\ 
  {...}&{...}&{...}&{...}&{...} \\ 
  {pos_1^m}&{pos_2^m}&{...}&{...}&{pos_n^m} 
\end{array}} \right]
\label{eq:2}
\end{equation}

Where $pos_i^m$ represents position $i^{th}$ UAV in $m$ dimensional space. Since a 3D space is used, therefore $M = 3$. The aim is to decrease path length $(pi_i)$ for each given UAV by following objective function:

\begin{equation}
(\pi _1^*,\,\pi _2^*,\,.\,.\,.\,,\,\pi _n^*) = \arg {\min _{{\pi _1},\,{\pi _2},\,.\,.\,.\,,\,{\pi _n}}}\sum\limits_{j = 1}^n {{c_j}{\pi _j}} 
\label{eq:3}
\end{equation}

subject to constraint

\begin{equation}
{\chi _{ij}}({\pi _i},\,{\pi _j}) = 0\,\,\,\,\,\,\,\,\,\,\forall \,i,j = 1,2,.\,.\,.,n
\label{eq:4}
\end{equation}

Where $i$ and $j$ represent the different UAVs; ${\pi_i}$ denotes the path of $i^{th}$ UAV which is computed by

\begin{equation}
{\pi _i} = \sqrt {{{({x_i} - {x_{t}})}^2} + {{({y_i} - {y_{t}})}^2} + {{({z_i} - {z_{t}})}^2}} 
\label{eq:5}
\end{equation}

Where, ${c_j}$ denotes cost associated with path ${\pi _j}$ \cite{ref-x}.

${\chi _{ij}}$ represents violation of the same path constraints between the trajectories of UAV $i$ and $j$. So $ i $ and $ j $ are taken from non-intersection paths from source to destination. Figure 2 shows the paths followed by the UAVs to reach their destinations while not suffering any collisions in between.
\begin{figure}[h]
\centering


\includegraphics[width=0.5\textwidth]{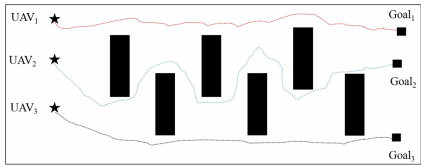}
\caption{Trajectory generated by the multiple UAVs in a 2D environment \cite{ref-t}}
\end{figure}


\subsection{Terrain construction}

To understand and solve the problem of route planning, an environment is required to simulate the UAVs to generate a route. The environment will contain several areas where the movement is prohibited, and those areas can be termed as obstacles. To prevent a collision, UAVs shall stay away from these areas. Obstacles of cuboidal shape with distinct sizes are chosen however in the real-life situations, obstacles generally don't possess a specific geometrical shape. To perform modeling in an environment where obstacles do not have a perfect geometrical shape is challenging and hampers the experimentation. Irregular obstacles are also included in modeling and testing of environment, and SSA will avert the uneven obstacles efficiently.

Two dimensional route planning is related to the planning of motion for vehicles on land and also in air or underwater while restricting one dimension. Simulation in 3-D environment is complex and requires expensive computations to generate the routes. At the same time, a 3D environment better models the complex real-life scenarios. 

\begin{figure}[h]
\centering
\subfigure[Map 1]{\includegraphics[width=55mm,scale=0.55]{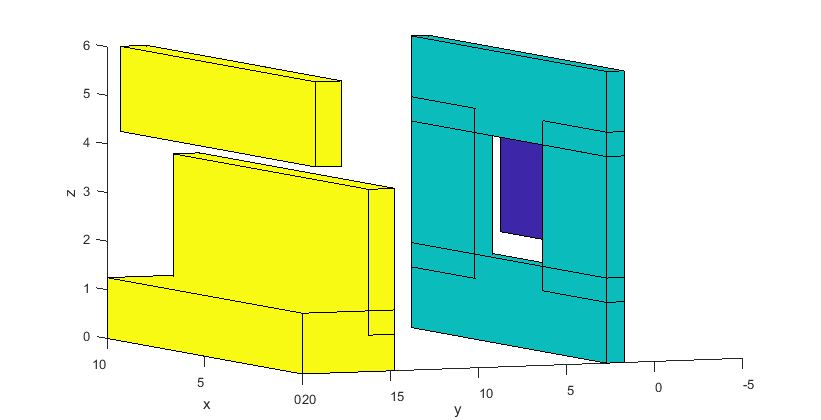}}
\hspace{1.0 mm}
\subfigure[Map 2]{\includegraphics[width=55mm,scale=0.55]{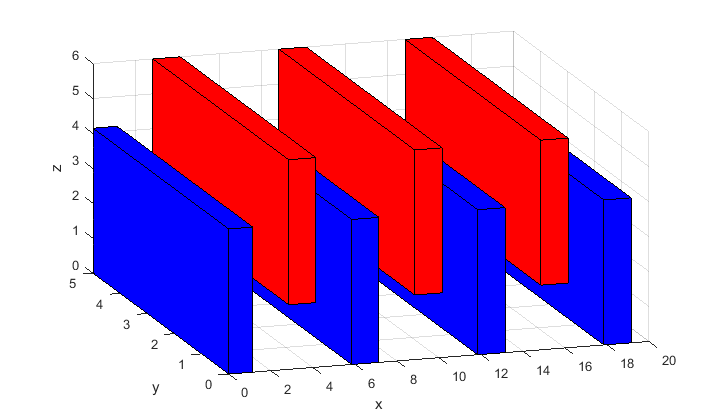}} \\
\subfigure[Map 3]{\includegraphics[width=55mm,scale=0.55]{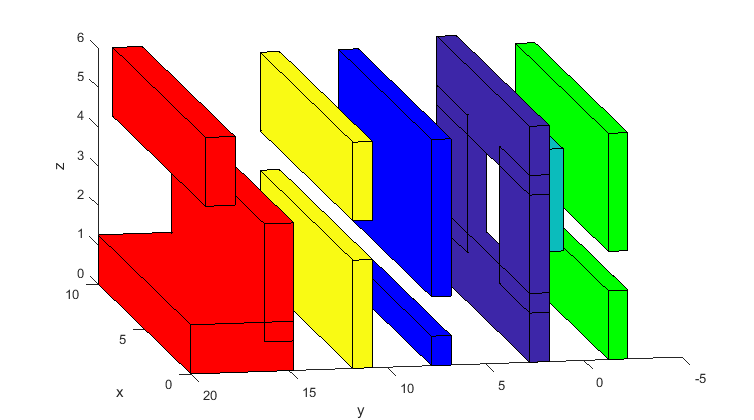}}
\hspace{1.0 mm}
\subfigure[Map 4]{\includegraphics[width=55mm,scale=0.55]{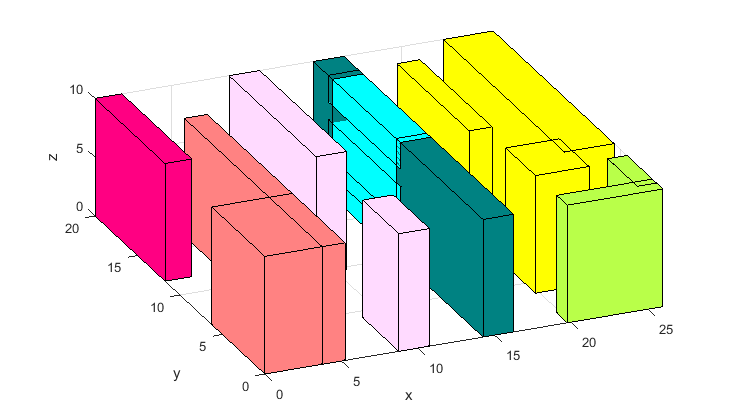}} 
\caption{Depiction of maps of 3D environment}
\label{fig:4}
\end{figure}

The details of the boundary, starting positions of UAVs and goal locations, as well as positions of obstacles for four maps used are depicted in Tables \ref{tab:1}, \ref{tab:2} and \ref{tab:3} respectively\cite{ref-m}. Using a different set of maps, as presented in Figure \ref{fig:4}, performance of SSA is compared and analyzed with deterministic and other meta-heuristic algorithms. 

\begin{table}[h]
\centering
\caption{3D Map Boundary Representation}
\label{tab:1}       
\begin{tabular}{lll}
\hline\noalign{\smallskip}
\textbf{Map} & \textbf{Start Boundary} & \textbf{End Boundary}  \\
\noalign{\smallskip}\hline\noalign{\smallskip}
Map 1 & (0, -5, 0) & (10, 20, 6) \\
Map 2 & (0, 0, 0) & (20, 5, 6) \\
Map 3 & (0, -5, 0) & (10, 20, 6) \\
Map 4 & (0, 0, 0) & (26, 20, 10)  \\
\noalign{\smallskip}\hline
\end{tabular}
\end{table}


\begin{table*}[h]
\centering
\caption{3D Map Start and Goal Representation}
\label{tab:2}       
{\begin{tabular*}{\textwidth}{@{\extracolsep{\fill}}lll}
\toprule
\textbf{Map} & \textbf{Start Position} & \textbf{Goal Position}  \\
\midrule
Map 1 & (2, 10, 2) (1, -4, 1) (9.2, 17, 3) (9.2, 10, 3) (0.1, 10, 2) & 
(1, -4, 1) (0.1, 17, 3) (9, -4, 1) (0.9, -4, 5) (9, 10, 2) \\
Map 2 & (0, 1, 5) (0, 2, 5) (0, 3, 5) (19, 4, 5) (19, 5, 5) & 
(19, 0, 5) (19, 5, 5) (19, 4, 5) (0, 3, 5) (0, 1, 5) \\
Map 3 & (2, 10, 2) (1, -4, 1) (9.2, 17, 3) (9.2, 10, 3) (0.1, 10, 2) & 
(1, -4, 1) (0.1, 17, 3) (9, -4, 1) (0.9, -4, 5) (9, 10, 2) \\
Map 4 & (17, 2, 5) (3, 2, 5) (7, 2, 5) (7, 18, 7) (17, 18, 5) & 
(21.5, 12, 4) (6, 6, 5) (13, 1, 3) (12, 14, 6) (22, 14, 8) \\
\botrule
\end{tabular*}}
\end{table*}



\begin{table*}[h]
\scriptsize
\centering
\caption{3D Map Obstacle representation}
\label{tab:3}       
{\begin{tabular*}{\textwidth}{@{\extracolsep{\fill}}lllll}
\toprule
\textbf{Obs}& \textbf{Map 1} & \textbf{Map 2} & \textbf{Map 3} & \textbf{Map 4}  \\
\midrule
1 & 
(0, 2, 4.5) - (10, 2.5, 6) & 
(3.1, 0, 2.1) - (3.9, 5, 6) & 
(0, -2, 0) - (10, -1.5, 1.5) & 
(0, 12, 0) - (1.5, 20, 10)\\

2 & 
(0, 2, 1.5) - (3, 2.5, 4.5) & 
(9.1, 0, 2.1) - (9.9, 5, 6) & 
(0,-2, 3) - (10, -1.5, 5.5) &
(0, 0, 0) - (5, 6, 10)\\

3 &
(7, 2, 1.5) - (10, 2.5, 4.5) &
(15.1, 0, 2.1) - (15.9, 5, 6) & 
(0, 2, 0) - (10, 2.5, 1.5) &
(4, 0, 0) - (5, 16, 10)\\

4 & (0, 2, 0) - (10, 2.5, 1.5) & 
(0.1, 0, 0) - (0.9, 5, 3.9) & 
(0, 2, 4.5) - (10, 2.5, 6) & 
(9, 10, 0) - (10.5, 20, 10)\\

5 & (0, 15, 0) - (10, 20, 1) & 
(6.1, 0, 0) - (6.9, 5, 3.9) & 
(0, 2, 1.5) - (3, 2.5, 4.5) &
(9, 0, 0) - (10.5, 4, 10)\\

6 &
(0, 15, 1) - (10, 16, 3.5) &
(12.1, 0, 0) - (12.9, 5, 3.9) & 
(7, 2, 1.5) - (10, 2.5, 4.5) &
(14.5, 0, 0) - (16, 10, 10)\\

7 &
(0, 18, 4.5) - (10, 19, 6) & 
(18.1, 0, 0) - (18.9, 5, 3.9) & 
(3, 0, 2.4) - (7, 0.5, 4.5) &
(14.5, 10, 8) - (16, 18, 10)\\

8 &
(3, 0, 2.4) - (7, 0.5, 4.5) &
- &
(0, 15, 0) - (10, 20, 1) &
(14.5, 10, 3.5) - (16, 18, 6)\\

9 & 
- & 
- & 
(0, 15, 1) - (10, 16, 3.5) &
(14.5, 10, 0) - (16, 18, 1.5)\\

10 &
- &
- &
(0, 18, 4.5) - (10, 19, 6) & 
(14.5, 18, 0) - (16, 20, 10)\\

11 &
- &
- &
(0, 7, 0) - (10, 7.5, 0.5) &
(19, 10, 0) - (20, 18, 10)\\

12 &
- &
- &
(0, 7, 2) - (10, 7.5, 5.5) &
(20, 4, 0) - (23, 7, 10)\\

13 &
- &
- &
(0, 11, 0) - (10, 11.5, 2.5) &
(20, 0, 0) - (26, 1, 10)\\

14 &
- &
- &
(0, 11, 4) - (10, 11.5, 5.5) &
(23, 6, 0) - (26, 20, 10)\\

15 &
- &
- &
- &
(25, 1, 0) - (26, 4, 10)\\

\botrule
\end{tabular*}}
\end{table*}


\subsection{Cost Function}

Due to the random nature of generated routes, there exists a possibility that the UAVs might collide with the obstacles and cannot continue its  motion, so the goal location cannot be reached. So, various costs are required to be introduced in the route planning problem like fuel cost, cost due to sharp turns and cost due to an incomplete route. The various costs involved can be illustrated as\cite{ref-m}:

\begin{itemize}
\item $C_{fuel}$: The cost of fuel which depends on length of the route followed. This cost is less when the route followed has a smaller length, which in turn leads to less consumption of fuel and lower time to reach the goal.
\item $C_{divergence}$: This cost is governed by frequency of sharp turns present in route. Lower the sharp turns the smoother and stable the route is.
\item $C_{gap}$: This cost corresponds to the separation among the goal and the route's end when the UAVs fail to reach the goal location. This situation could arise due to a significant obstacle which lies in between the route to goal. This cost corresponds to the highest interest amongst all the costs and assigned with the highest priority. This cost should have a zero value in the optimal solution.
\end{itemize}

The generated route $R_{(x,y,z)}$ to the goal location can be presented as a sequence of points from the source location as follows: 

\begin{equation}
R_{x,y,z} = \{S, a_1, a_2, a_3, a_4, ...a_n, G\}     
\end{equation}

where, $S$ represents the start location; $G$ represents the goal location; $a_1,a_2, ...a_n$ represent the points occurring in the route.

To determine the cost of the fuel, the speed of the UAV is assumed to be constant during route planning. Thus, the cost of the fuel can be obtained by using total distance traveled by the UAVs. The cost of fuel is presented using Equation \ref{eqn:sum1}.

\begin{equation}
C_{fuel} = \sum (D_{x, y, z})
\label{eqn:sum1}
\end{equation}

The cost due to sharp turns occurs when change in direction from $a_{j-1}$ to $a_j$ do not match the change in direction from $a_j$ to $a_{j+1}$. It can be computed by searching the cases in obtained route with a considerable amount of turn in the angle between the two. This cost will increase with the increase of sharp turns in the planned route. This cost can be inferred from equation \ref{eqn:sum2}.

\begin{equation}
C_{divergence} = \textrm{number of turning points in route}
\label{eqn:sum2}
\end{equation}

Now the most significant cost, i.e., $C_{gap}$ is computed using the Euclidean distance. The two required points are end point of route and the goal location. If the endpoint and goal location is the same than this cost is set to 0 else, it is determined with the help of equation \ref{eqn:sum3}.

\begin{equation}
    c_{gap} = \sqrt{(X_{end} - T_x)^2 + (Y_{end} - T_y)^2 + (Z_{end} - T_z)^2}
    \label{eqn:sum3}
\end{equation}

$(X_{end}, Y_{end}, Z_{end})$ represent the endpoints of the generated route; $(T_x, T_y, T_z)$ represents the target location.

The total cost of a route can be computed using equation \ref{eqn:sum4}.

\begin{equation}
    C_{total} = P_{1} \times C_{fuel} + P_{2} \times C_{divergence} + P_{3} \times C_{gap}
    \label{eqn:sum4}
\end{equation}

The total cost function helps to achieve the optimal route from the route set.

Here $P_{1}, P_{2}, P_{3}$ are the experimental parameters and their values are actuated with the help of experiments and constraints of the problem description. In scenarios where a more substantial route length is taken as a preference, then $P_{1}$ is assigned a higher value. Similarly $P_{2}$ assigns inclination towards the route smoothness. From $P_{3}$, it is ensured that the route planned reaches the goal. The UAVs posses a propensity to include information regarding their locality only, and for the subsequent increment, they might alter their location among one of the cells in the neighborhood. Thus, UAVs do not have any prior knowledge regarding at what time and location it will face an obstacle; that is why UAVs steadily continue to modify their paths whenever they encounter the obstacles.


\subsection{Trajectory generation for multiple UAVs}

The generated trajectory by the SSA may include some sharp turns which practically are not feasible for UAVs to follow. So, it becomes crucial to smoothen the generated route. The movement and speed of the UAVs can be presented using polynomial functions for vehicular mobility. The polynomial function involves the component of time. A continuous route is generated by the fifth-degree polynomial, which indicates that the first derivatives are consistent. The polynomial derivative can be determined efficiently to obtain the results of the campaign in this manner. A conventional fifth order time polynomial function\cite{ref-x} is represented using equation \ref{eqn:sum5}.

\begin{equation}
   S(t) = a_{5}t^5 + a_{4}t^4 + a_{3}t^3 + a_{2}t^2 + a_{1}t^1 + a_{0}
   \label{eqn:sum5}
\end{equation}

While generating the trajectory, the sub-points are obtained in the initial phase. Sub-point can be considered as a necessary point in the confined route, which assists the UAV to avert any obstacle or steep turn. To produce a smooth route, the speed and movement of UAV traveling through sub-points are utilized by the fifth order polynomial based trajectory fitting strategy\cite{ref-x}. 

The main objective of using SSA for multiple UAV route planning in a 3D environment is:

\begin{itemize}
    \item to localize the UAVs and destination location.
    \item to plan a route between multiple UAVs and their corresponding targets and simultaneously ensure that no UAV collides with each other or any obstacle.
\end{itemize}

\section{Salp Swarm Algorithm}
\label{sec:4}

\begin{figure}[h]
\centering
\subfigure[Single Salp] {\includegraphics[width=20mm,scale=0.20]{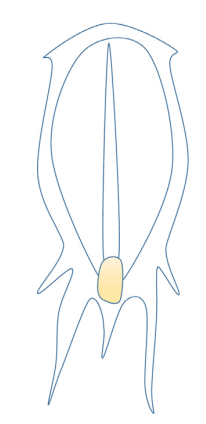}}
\hspace{19.0 mm}
\subfigure[Salp Chain] {\includegraphics[width=40mm,scale=0.40]{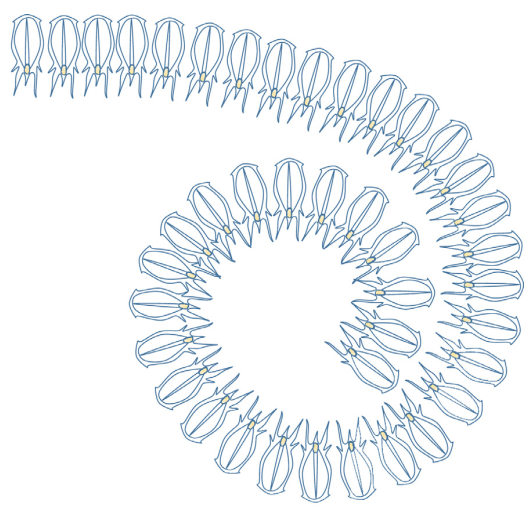}} 
\caption{Schematic Diagram of Salps \cite{ref-s}}
\label{fig:5}
\end{figure}

Salp is a sea creature having a transparent barrel-shaped body. It is a part of the \textit{Salpidae} family. It's movement is governed by pumping of water through the body and as propulsion to move forward. SSA is inspired by the swarming behavior of salps while foraging and navigating in the water. The
shape of a salp is shown in Figure \ref{fig:5}(a). The swarm formed by the salps in deep oceans is referred to as a salp chain. This chain is illustrated in Figure \ref{fig:5}(b). This algorithm was first developed by Mirjalili et. al. in 2017\cite{ref-s}. The primary reason to choose SSA for route planning was its simplicity and since the inspiration for the algorithm is from the natural navigating and foraging behavior of the salps in the ocean in search of food. With respect to multi-UAV route planning, the foraging phase of the salp chains can be considered as a search for different targets in the environment and the navigating phase can be regarded as connecting the points in the environment to obtain waypoints for the generated path length.

For understanding the salp chains, a mathematical model was developed. The population is broken down into 2 segments- leader and followers. Leader salp is present at the front of the chain while remaining salps are recognized as followers. The salp at the front escorts the swarm while the remaining salps follow each other and leader (either directly or indirectly).
%

Similarly to other swarm-based techniques, the position of salps is defined in an $n$-dimensional search space where $n$ denotes number of variables for a given problem. Thus, the position of all salps is saved in a two-dimensional matrix called $x$. It is an assumption that there exists a food source called $F$ in the search space as the swarms' target. 

The complete working of SSA can be observed from the following equations.

Equation \ref{eqn:sum6} updates the position of the leader:

\begin{equation}
 x^{1}_{j} = \left\{ {\begin{array}{*{20}{l}}
  F_j + c_1 ((ub_j - lb_j)c_2 + lb_j) & {c_3 \geq 0} \\
  F_j - c_1 ((ub_j - lb_j)c_2 + lb_j) & {c_3 < 0} \\
\end{array}} \right.
\label{eqn:sum6}
\end{equation}

Where $x^1_j$ = position of the leader; $F_j$ = position of the food source; $ub_j$ and $lb_j$ are upper and lower bounds in $j^{th}$ dimension; $c_1$, $c_2$, $c_3$ are uniform random numbers. 

In Equation \ref{eqn:sum7}, $c_1$ maintains a balance between exploration \& exploitation phase:

\begin{equation}
    c_1 = 2 e^ {-(\frac {4l} {L}) ^ 2}
    \label{eqn:sum7}
\end{equation}

Where, $l$ : current iteration, $L$ : maximum number of iterations. 

Exploration is defined as the phase in which the algorithm tries to explore the search space. The avoidance of local solutions takes place in this phase. After the exploration comes the exploitation phase in which the primary concern is to improvise the solutions explored in the exploration phase.

Equation \ref{eqn:sum8} updates position of salps except for the leader:

\begin{equation}
    x_j^i = \frac{1}{2} (x_j^i + x^{i-1}_j)
    \label{eqn:sum8}
\end{equation}

The pseudocode of SSA algorithms is given in Algorithm \ref{a:ssa}.

\begin{algorithm}[H]
\caption{SSA Algorithm}
\label{a:ssa}
\begin{algorithmic}[1]

\STATE Initialize the salp population $x_i (i = 1, 2, ..., n)$ considering $ub$ and $lb$ 
		\STATE \textbf{while} (end condition is not satisfied)
		\STATE Calculate the fitness of each search agent (salp)
		\STATE $F$ = the best search agent
		\STATE Update $c_1$ by Eq. (2)
		\STATE \hspace*{0.5cm} \textbf{for} each salp ($x_i$)
		\STATE \hspace*{1cm} \textbf{if} ($i$ == 1) 
		\STATE \hspace*{1.5cm} Update the position of the leading salp by Eq. (1)
		\STATE \hspace*{1cm} \textbf{else}
		\STATE \hspace*{1.5cm} Update the position of the follower salp by Eq. (4)
		\STATE \hspace*{1cm} \textbf{end}
		\STATE \hspace*{0.5cm} \textbf{end}
		\STATE \hspace*{0.5cm} Amend the salps based on the upper and lower bounds of variables
		\STATE \textbf{end}
		\STATE return $F$

\end{algorithmic}
\end{algorithm}


Some assumptions are made in the research to get more efficient and useful results. The assumptions made here are as follows: 

\begin{itemize}
    \item The UAV is considered as a point object. 
    \item The speed of the UAV is kept constant during the entire simulation.  
    \item Upon reaching the target point, a UAV can stop immediately irrespective of the momentum.
\end{itemize}

SSA algorithm has computational complexity $O(t(d \times n + Cof \times n))$
\\ where $t$ = number of iterations; $d$ = number of variables (dimension); $n$ = number of solutions; and $Cof$ = the cost of the objective function.

\section{Implementation Procedure}
\label{sec:5}

To validate the effectiveness of SSA on multiple UAV route planning problem in a static 3D environment, a set of experiments have been performed in MATLAB. Matlab 2017a version and a PC with intel $i7$ processor, 3.40 GHz of CPU and 8 GB of RAM were chosen for performing the experimentation. In this section, experimental simulation is described, which is divided into two experiments. Then convergence analysis of the SSA algorithm on different maps is presented. 


\subsection{Simulation}

Before analyzing the performance of SSA in a 3D environment for multi-UAV route planning, it is first analyzed in a 2D environment and its performance as compared to deterministic and other meta-heuristic algorithms, is examined. To illustrate the above experiments, a different set of maps corresponding to a different environment have been taken.

\subsubsection{Experiment 1}

The performance of different algorithms is investigated in a 2D environment whose layout is given in Table \ref{tab:4}.


\begin{table}[h]
\centering
\caption{Static 2D Environment Layout\cite{ref-m}}
\begin{tabular}{c c c}
\hline\noalign{\smallskip}
\textbf{Obs No} & \textbf{Center (unit)} & \textbf{Radius (units)} \\
\noalign{\smallskip}\hline\noalign{\smallskip}
1 & (1.5, 4.5) & 1.5 \\
2 & (4.0, 3.0) & 1.0 \\
3 & (1.2, 1.5) & 0.8 \\
\noalign{\smallskip}\hline\noalign{\smallskip}
\multicolumn{3}{c}{Map size: 30 $\times$ 30 i.e. (--10 to +20)} \\
\noalign{\smallskip}\hline
\end{tabular}
\label{tab:4}
\end{table}


The initial parameters and constants for the algorithms used in experimentation are listed in Table \ref{tab:5}.


\begin{table}[h]
\centering
\caption{Parameter Setting for the SSA\cite{ref-t}}
\label{tab:5}
\begin{tabular}{c c c}
\noalign{\smallskip}\hline
\textbf{Algorithm} & \textbf{Parameters} & \textbf{Values }\\
\noalign{\smallskip}\hline\noalign{\smallskip}
\multirow{3}{*}{SSA} & Random Number 1 ($c_1$) & [0, 1] \\
 & Random Number 2 ($c_2$) & [0, 1] \\
 & Random Number 3 ($c_3$) & [0, 1] \\
\noalign{\smallskip}\hline
\end{tabular}
\end{table}
 
\begin{figure}[h]
\begin{minipage}{.5\linewidth}
\centering
\subfigure[Initial path generated during application of SSA for 2D environment.]{\label{main:a}\includegraphics[scale=0.20]{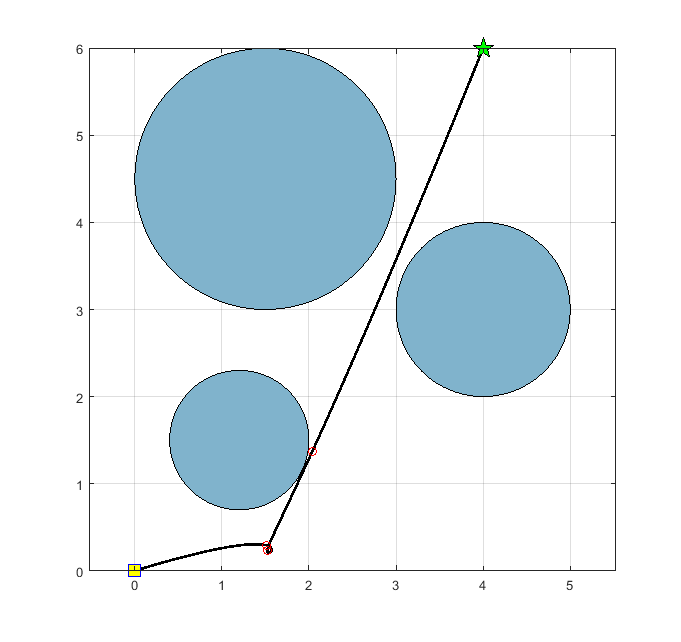}}
\end{minipage}%
\begin{minipage}{.5\linewidth}
\centering
\subfigure[Final obtained path]{\label{main:b}\includegraphics[scale=0.25]{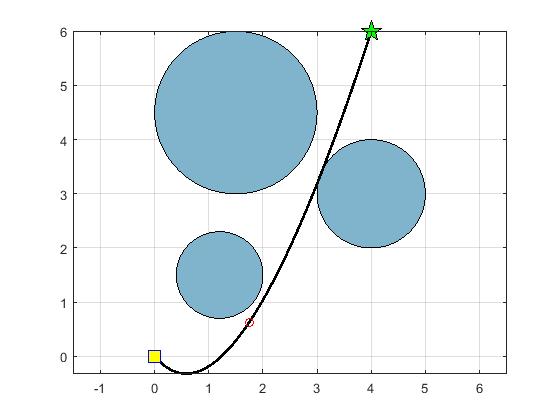}}
\end{minipage}\par\medskip
\centering
\subfigure[Best cost vs number of iteration graph]{\label{main:c}\includegraphics[scale=0.25]{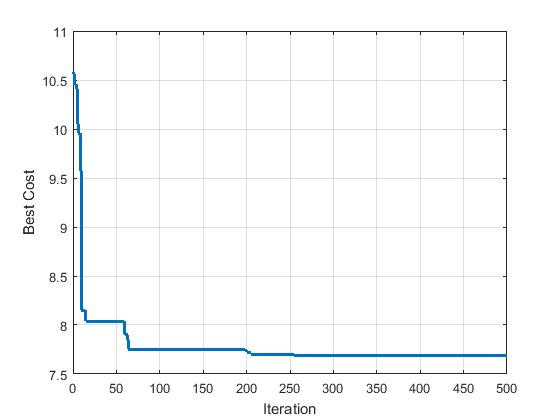}}
\caption{Simulation of SSA in 2D environment}
\label{fig:6a}
\end{figure}

The circular shape in the environment can be modeled as obstacles which should be completely avoided by the vehicle to prevent a collision. The start location of the vehicle was taken as (0, 0) and the goal location as (4, 6) \cite{ref-m}. The initial experiments of SSA for route planning in 2D environment are performed to test the effectiveness of the algorithm implementation and to have a conjecture on the quality of solution for the 3D path planning of UAV. The convergence curve and planned routes for SSA are illustrated in Figure \ref{fig:6a}.


The Figure \ref{fig:6a} depicts convergence analysis of SSA in the 2D arena simulated above. From Figure 5(c) it is clear that SSA converges quickly and there is little change in the cost with further increase in the number of iterations. The results of different algorithms for  with respect to the cost  and time in the 2D arena for single bot route planning in static environment are presented in Table \ref{tab:6}. 


\begin{table}[H]
\scriptsize
\centering
\caption{Results in static 2D environment}
\label{tab:6}
\begin{tabular}{l c c c c}
\noalign{\smallskip}\hline
\textbf{Algorithm} & \textbf{Population} & \textbf{Iteration} & \textbf{Best Cost (points)} & \textbf{Time (units)} \\
\noalign{\smallskip}\hline 
WOA & 150 & 500 & 8.0131 & 194.9152 \\
SCA & 150 & 500 & 8.0042 & 197.0136 \\
GSO & 150 & 500 & 8.0236 & 196.3792 \\
PSO & 150 & 500 & 8.0234 & 196.3822 \\
IBA & 150 & 500 & 7.9321 & 192.8920 \\
BBO & 150 & 500 & 7.9803 & 191.2722 \\
GWO & 150 & 500 & 7.9560 & 189.4108 \\
\textbf{SSA} & \textbf{150} & \textbf{500} & \textbf{7.9340} & \textbf{186.4025} \\
\noalign{\smallskip}\hline
\end{tabular}
\end{table}


%
%

\subsubsection{Experiment 2}

After analyzing the performance of SSA in a 2D environment, simulations are performed in a 3D environment. A set of four different maps are taken whose layouts are depicted in Table \ref{tab:3}.

The start and the end point concerning different obstacle number show the location of two diagonally opposite corners of the cuboid. Every map has different dimensions, different start, goal locations along with different lower and upper bound as given in Tables \ref{tab:1} and \ref{tab:2}.

\begin{figure}[h]
\centering
\subfigure[Map 1]{\includegraphics[width=65mm,scale=0.55]{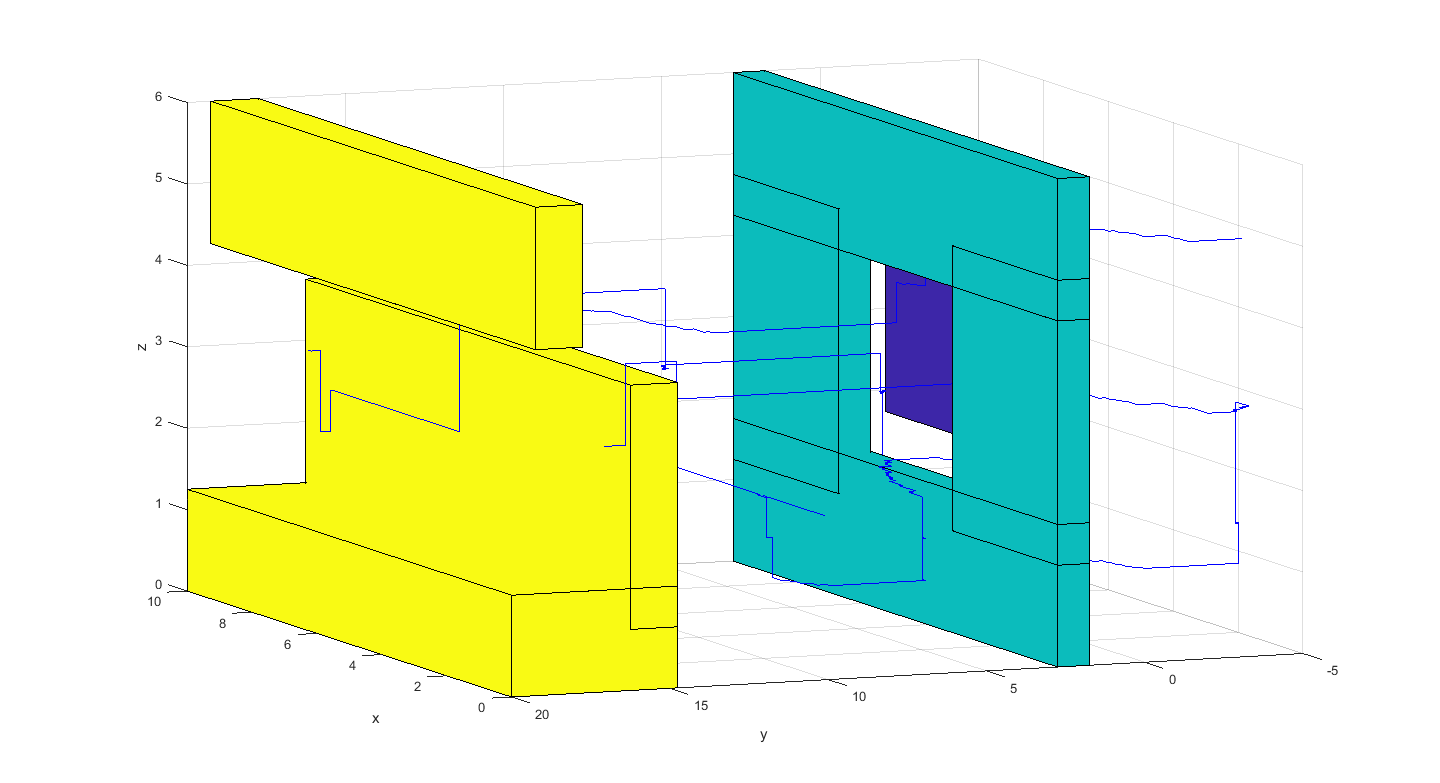}}
\hspace{1.0 mm}
\subfigure[Map 2]{\includegraphics[width=65mm,scale=0.55]{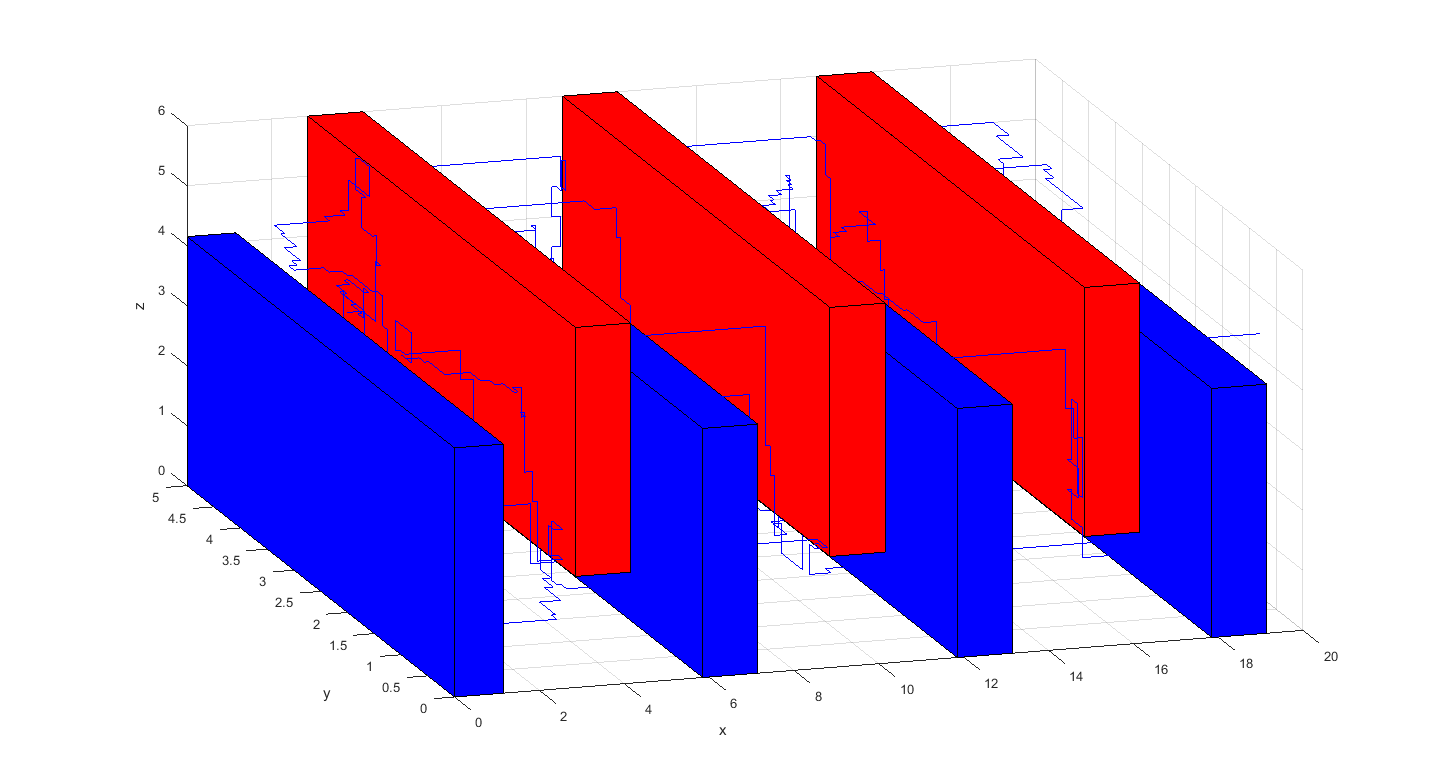}} \\
\subfigure[Map 3]{\includegraphics[width=65mm,scale=0.55]{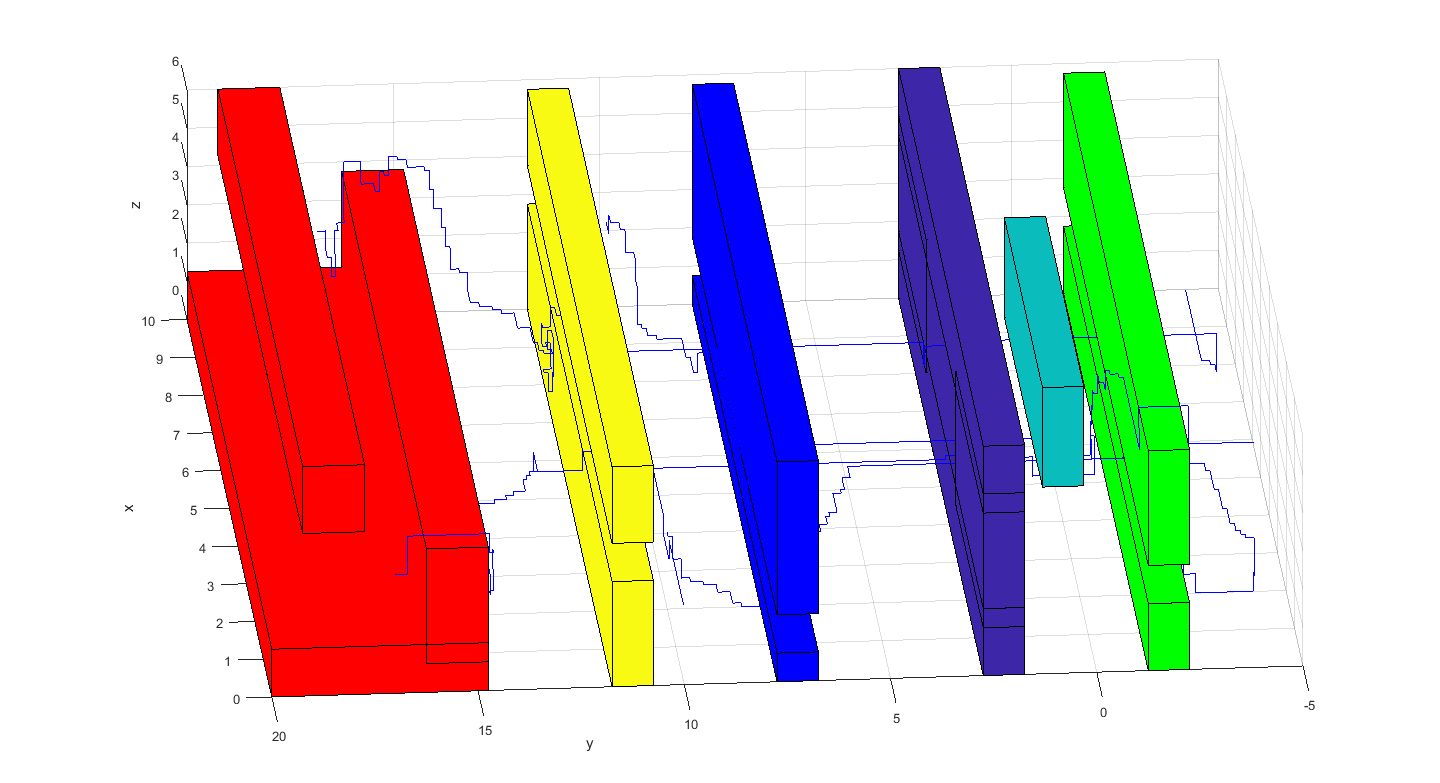}}
\hspace{1.0 mm}
\subfigure[Map 4]{\includegraphics[width=65mm,scale=0.55]{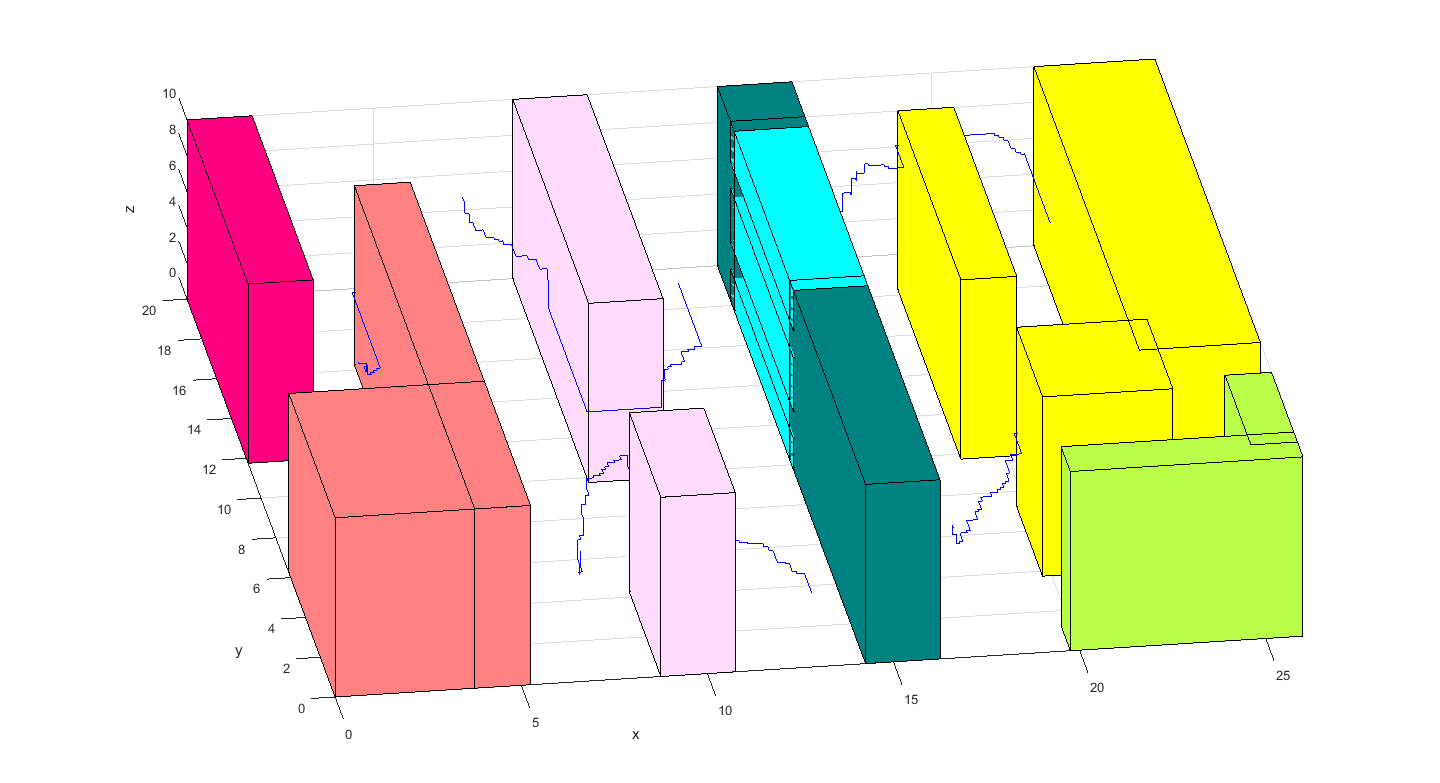}} 
\caption{Simulation of SSA on Maps}
\label{fig:7-x}
\end{figure}

The initial parameters and constants for different algorithms are the same as in the case of a 2D  environment and are given in Table \ref{tab:5}. Figure \ref{fig:7-x} shows the trajectory generated by SSA on various maps.

For map 1, 2 and 3 all the algorithms listed above were able to find the paths without any collisions. Map 4 is relatively complex and not all the algorithms discovered collision-free path in first run. SSA performed satisfactorily in map 4 and discovered a collision-free path in all of the multiple runs. To compare performance of different algorithms wrt time and cost, Tables \ref{tab:12} - \ref{tab:10.3} are formulated.

\begin{table*}
\centering
\caption{Execution time and best cost of algorithms for Map 1 for different population size and max iterations}
\label{tab:12}
{\smallskip
{\begin{tabular*}{\textwidth}{@{\extracolsep{\fill}}ccccccccc}
\toprule
Algorithm  & Pop. Size & Iterations  & \begin{tabular}[c]{@{}c@{}}Best Cost\\  $(UAV_1)$\\(cm)\end{tabular} & \begin{tabular}[c]{@{}c@{}}Best Cost \\ $(UAV_2)$\\(cm)\end{tabular} & \begin{tabular}[c]{@{}c@{}}Best Cost \\ $(UAV_3)$\\(cm)\end{tabular} & \begin{tabular}[c]{@{}c@{}}Best Cost \\ $(UAV_4)$\\(cm)\end{tabular} & \begin{tabular}[c]{@{}c@{}}Best Cost \\ $(UAV_5)$\\(cm)\end{tabular} & \begin{tabular}[c]{@{}c@{}}Overall Time \\(sec)\end{tabular}    \\
\midrule 
WOA & 20 & 25 & 215 & 310 & 283 & 232 & 92 & 57.64\\ 
SCA & 20 & 25 & 199 & 306 & 293 & 232 & 92 & 59.90\\
GSO & 20 & 25 & 221 & 296 & 283 & 236 & 92 & 57.50\\
PSO & 20 & 25 & 221 & 296 & 283 & 232 & 92 & 57.15\\ 
IBA & 20 & 25 & 215 & 308 & 279 & 232 & 92 & 72.15\\  
BBO & 20 & 25 & 201 & 322 & 299 & 232 & 92 & 65.70\\ 
GWO & 20 & 25 & 201 & 284 & 295 & 232 & 92 & 48.50\\ 
\textbf{SSA} & \textbf{20} & \textbf{25} & \textbf{202} & \textbf{284} & \textbf{290} & \textbf{230} & \textbf{92} & \textbf{45.30} \\ 
WOA & 25 & 40 & 199 & 308 & 285 & 228 & 92 & 65.75\\ 
SCA & 25 & 40 & 213 & 296 & 285 & 232 & 94 & 67.35\\
GSO & 25 & 40 & 201 & 284 & 291 & 230 & 96 & 63.49\\
PSO & 25 & 40 & 207 & 298 & 285 & 232 & 94 & 91.27\\ 
IBA & 25 & 40 & 208 & 300 & 279 & 230 & 92 & 92.32\\  
BBO & 25 & 40 & 201 & 300 & 289 & 232 & 92 & 70.40\\  
GWO & 25 & 40 & 201 & 284 & 295 & 232 & 92 & 56.98\\  
\textbf{SSA} & \textbf{25} & \textbf{40} & \textbf{200} & \textbf{280} & \textbf{285} & \textbf{230} & \textbf{92} & \textbf{50.03} \\
\botrule
\end{tabular*}}
}
\end{table*}


\begin{table*}
\centering
\caption{Execution time and best cost of algorithms for Map 2 for different population size and max iterations}
\label{tab:10.1}
{\smallskip
{\begin{tabular*}{\textwidth}{@{\extracolsep{\fill}}ccccccccc}
\toprule
Algorithm  & Pop. Size & Iterations  & \begin{tabular}[c]{@{}c@{}}Best Cost\\  $(UAV_1)$\\(cm)\end{tabular} & \begin{tabular}[c]{@{}c@{}}Best Cost \\ $(UAV_2)$\\(cm)\end{tabular} & \begin{tabular}[c]{@{}c@{}}Best Cost \\ $(UAV_3)$\\(cm)\end{tabular} & \begin{tabular}[c]{@{}c@{}}Best Cost \\ $(UAV_4)$\\(cm)\end{tabular} & \begin{tabular}[c]{@{}c@{}}Best Cost \\ $(UAV_5)$\\(cm)\end{tabular} & \begin{tabular}[c]{@{}c@{}}Overall Time \\(sec)\end{tabular}    \\ 
\midrule
WOA	& 20 & 25 & 345 & 333 & 323 & 417 & 341	& 145.67\\
SCA	& 20 & 25 & 333 & 323 & 343 & 385 & 455	& 142.90\\
GSO & 20 & 25 & 355 & 325 & 327 & 387 & 391 & 136.27\\ 
PSO & 20 & 25 & 333 & 313 & 337 & 365 & 423 & 136.06\\ 
IBA & 20 & 25 & 313 & 381 & 365 & 393 & 401 & 138.54\\ 
BBO & 20 & 25 & 375 & 405 & 629 & 758 & 389 & 256.87\\
GWO & 20 & 25 & 321 & 381 & 317 & 353 & 375 & 132.90\\
\textbf{SSA} & \textbf{20} & \textbf{25} & \textbf{321} & \textbf{378} & \textbf{313} & \textbf{352} & \textbf{375} & \textbf{125.76}\\
\hline
WOA	& 30 & 40 & 313	& 311 & 371 & 367 & 403	& 152.20\\
SCA	& 30 & 40 & 357 & 341 & 351 & 403 & 373 & 153.87\\
GSO & 30 & 40 & 339 & 321 & 317 & 333 & 319 & 146.29\\ 
PSO & 30 & 40 & 325 & 343 & 299 & 327 & 377 & 197.94\\ 
IBA & 30 & 40 & 331 & 343 & 397 & 341 & 355 & 146.01\\ 
BBO & 30 & 40 & 387 & 375 & 345 & 704 & 389 & 266.83\\
GWO & 30 & 40 & 319 & 313 & 315 & 333 & 345 & 142.90\\
\textbf{SSA} & \textbf{30} & \textbf{40} & \textbf{318} & \textbf{305} & \textbf{312} & \textbf{329} & \textbf{342} & \textbf{130.87}\\
\botrule
\end{tabular*}}
}
\end{table*}


\begin{table*}
\centering
\caption{Execution time and best cost of algorithms for Map 3 for different population size and max iterations}
\label{tab:10.2}

{\begin{tabular*}{\textwidth}{@{\extracolsep{\fill}}ccccccccc}
\toprule
Algorithm  & Pop. Size & Iterations  & \begin{tabular}[c]{@{}c@{}}Best Cost\\  $(UAV_1)$\\(cm)\end{tabular} & \begin{tabular}[c]{@{}c@{}}Best Cost \\ $(UAV_2)$\\(cm)\end{tabular} & \begin{tabular}[c]{@{}c@{}}Best Cost \\ $(UAV_3)$\\(cm)\end{tabular} & \begin{tabular}[c]{@{}c@{}}Best Cost \\ $(UAV_4)$\\(cm)\end{tabular} & \begin{tabular}[c]{@{}c@{}}Best Cost \\ $(UAV_5)$\\(cm)\end{tabular} & \begin{tabular}[c]{@{}c@{}}Overall Time \\(sec)\end{tabular}    \\ 
\midrule
WOA	& 20 & 25 & 219 & 328 & 343 & 256 & 94 & 95.02\\
SCA	& 20 & 25 & 235 & 302 & 393 & 278 & 92 & 99.75\\
GSO & 20 & 25 & 219 & 322 & 339 & 294 & 92 & 108.69\\ 
PSO & 20 & 25 & 231 & 324 & 371 & 244 & 94 & 95.76\\ 
IBA & 20 & 25 & 203 & 320 & 425 & 268 & 92 & 102.47\\ 
BBO & 20 & 25 & 215 & 346 & 716 & 266 & 94 & 108.74\\
GWO & 20 & 25 & 213 & 300 & 355 & 274 & 94 & 92.46\\
\textbf{SSA} & \textbf{20} & \textbf{25} & \textbf{210} & \textbf{300} & \textbf{345} & \textbf{274} & \textbf{94} & \textbf{90.30}\\
\hline
WOA	& 25 & 40 & 245	& 308 & 347 & 246 & 96 & 101.81\\
SCA	& 25 & 40 & 225 & 280 & 369 & 246 & 92 & 105.79\\
GSO & 25 & 40 & 215 & 324 & 335 & 254 & 92 & 141.91\\ 
PSO & 25 & 40 & 215 & 318 & 333 & 254 & 92 & 146.95\\ 
IBA & 25 & 40 & 247 & 324 & 375 & 302 & 92 & 151.41\\ 
BBO & 25 & 40 & 215 & 346 & 716 & 266 & 94 & 108.74\\
GWO & 25 & 40 & 201 & 312 & 339 & 254 & 92 & 104.87\\
\textbf{SSA} & \textbf{25} & \textbf{40} & \textbf{202} & \textbf{305} & \textbf{340} & \textbf{250} & \textbf{91} & \textbf{102.55}\\
\botrule
\end{tabular*}}
\end{table*}


\begin{table*}
\centering
\caption{Execution time and best cost of algorithms for Map 4 for different population size and max iterations}
\label{tab:10.3}

{\begin{tabular*}{\textwidth}{@{\extracolsep{\fill}}ccccccccc}
\toprule 
Algorithm  & Pop. Size & Iterations  & \begin{tabular}[c]{@{}c@{}}Best Cost\\  $(UAV_1)$\\(cm)\end{tabular} & \begin{tabular}[c]{@{}c@{}}Best Cost \\ $(UAV_2)$\\(cm)\end{tabular} & \begin{tabular}[c]{@{}c@{}}Best Cost \\ $(UAV_3)$\\(cm)\end{tabular} & \begin{tabular}[c]{@{}c@{}}Best Cost \\ $(UAV_4)$\\(cm)\end{tabular} & \begin{tabular}[c]{@{}c@{}}Best Cost \\ $(UAV_5)$\\(cm)\end{tabular} & \begin{tabular}[c]{@{}c@{}}Overall Time \\(sec)\end{tabular}    \\ 
\midrule
WOA	& 20 & 25 & 256 & 388 & 459 & 318 & 122 & 229.66 \\
SCA	& 20 & 25 & 259 & 383 & 457 & 320 & 122 & 211.86 \\
GSO & 20 & 25 & 249 & 352 & 369 & 324 & 122 & 141.91\\
PSO & 20 & 25 & 261 & 354 & 401 & 274 & 124 & 95.76\\
IBA & 20 & 25 & 233 & 350 & 455 & 298 & 122 & 102.47\\
BBO & 20 & 25 & 245 & 376 & 746 & 296 & 124 & 274.43\\
GWO & 20 & 25 & 243 & 330 & 385 & 304 & 124 & 92.46\\
\textbf{SSA} & \textbf{20} & \textbf{25} & \textbf{242} & \textbf{325} & \textbf{385} & \textbf{281} & \textbf{124} & \textbf{85.20}\\ 
\hline
WOA	& 25 & 40 & 256 & 365 & 359 & 309 & 122	& 247.13\\
SCA	& 25 & 40 & 260 & 374 & 358	& 308 & 122 & 211.42\\
GSO & 25 & 40 & 245 & 354 & 365 & 284 & 122 & 108.69\\
PSO & 25 & 40 & 245 & 348 & 363 & 272 & 122 & 146.95\\
IBA & 25 & 40 & 277 & 354 & 405 & 332 & 122 & 151.41\\
BBO & 25 & 40 & 231 & 350 & 355 & 302 & 122 & 108.74\\
GWO & 25 & 40 & 255 & 342 & 369 & 284 & 122 & 104.87\\
\textbf{SSA} & \textbf{25} & \textbf{40} & \textbf{250} & \textbf{322} & \textbf{370} & \textbf{276} & \textbf{122} & \textbf{101.44}\\
\botrule
\end{tabular*}}
\end{table*}



\begin{figure}[h]
\centering
\subfigure[Time taken for different algorithms]{\includegraphics[width=73mm,scale=0.7]{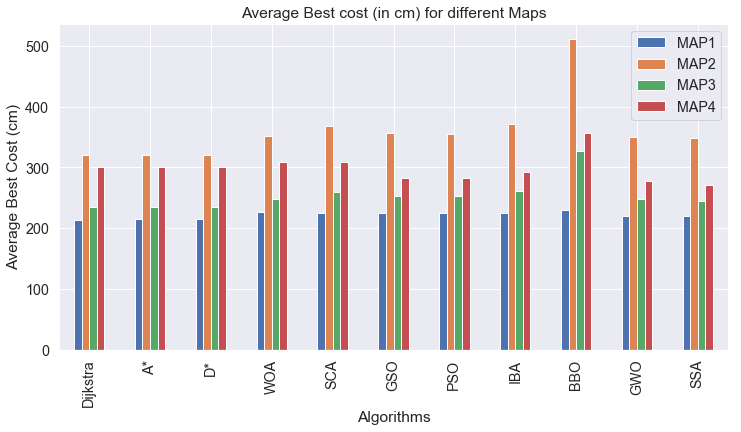}}
\hspace{5.0 mm}
\subfigure[Avg best cost for different algorithms]{\includegraphics[width=73mm,scale=0.7]{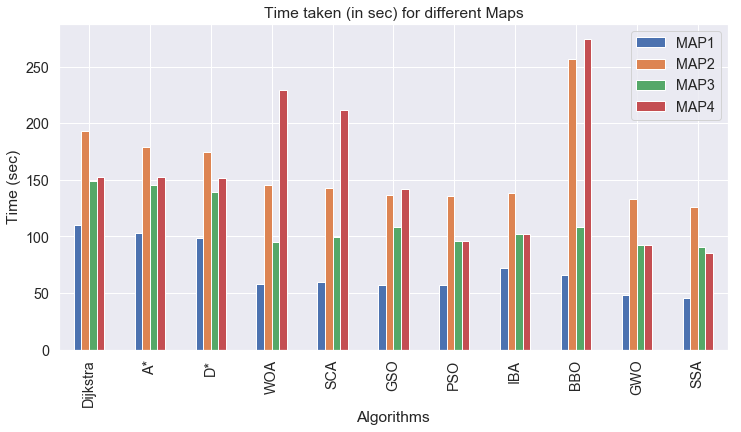}} 
\caption{Time and Cost Graphs for Different Algorithms}
\label{fig:8}
\end{figure}

The results from the tables can be visualized in Figure \ref{fig:8}.

\subsection{Convergence analysis of SSA algorithm}
\label{section-5.2}
By analyzing the following Tables \ref{tab:13}-\ref{tab:16}, it becomes evident that the SSA algorithm generates near-optimal results in significantly fewer iterations if there are no obstacles in the workspace. A number of iterations are required to converge towards the optima when obstacles are presented in the workspace; but after completing 15-20 iterations, results do not alter even if more increments are introduced.


\begin{table}[h]
\scriptsize
\centering
\caption{Average best cost (in cm) per iteration count for Map 1}
\begin{tabular}{c c c c c c c c c}
\hline
Iteration & GWO & GSO & PSO & BBO & IBA & WOA & SCA & \textbf{SSA}\\
\hline
1 & 250.80 & 260.80 & 252.00 & 419.79 & 328.40 & 256.40 & 267.20 & \textbf{256.65}\\
5 & 238.40 & 245.20 & 238.80 & 264.80 & 249.60 & 239.20 & 238.80 & \textbf{235.55}\\
10 & 228.80 & 230.00 & 236.80 & 243.69 & 238.40 & 229.40 & 236.20 & \textbf{224.80}\\
15 & 222.80 & 228.40 & 228.80 & 242.40 & 237.20 & 224.40 & 226.80 & \textbf{219.60}\\
20 & 221.20 & 227.20 & 228.60 & 232.00 & 234.60 & 222.40 & 225.20 & \textbf{218.65}\\
25 & 220.80 & 226.80 & 220.80 & 226.40 & 234.00 & 221.60 & 222.80 & \textbf{218.05}\\
30 & 219.60 & 224.20 & 224.40 & 226.00 & 228.80 & 219.80 & 221.60 & \textbf{217.80}\\
35 & 219.20 & 222.40 & 224.00 & 224.40 & 226.40 & 219.20 & 219.20 & \textbf{217.40}\\
\hline
\end{tabular}
\label{tab:13}
\end{table}


\begin{table}[h]
\scriptsize
\centering
\caption{Average best cost (in cm) per iteration count for Map 2}
\begin{tabular}{c c c c c c c c c}
\hline
Iteration & GWO & GSO & PSO & BBO & IBA & WOA & SCA & \textbf{SSA}\\
\hline
1 & 644.03 & 644.88 & 698.75 & 698.56 & 1038.79 & 690.40 & 696.60 & \textbf{621.92}\\
5 & 392.80 & 394.26 & 425.67 & 474.20 & 683.41 & 396.00 & 398.20 & \textbf{440.76}\\
10 & 385.40 & 391.40 & 415.60 & 422.60 & 512.52 & 386.60 & 388.80 & \textbf{382.50}\\
15 & 361.00 & 390.40 & 398.60 & 385.40 & 507.04 & 374.60 & 373.80 & \textbf{347.81}\\
20 & 355.00 & 390.20 & 394.40 & 379.80 & 452.02 & 360.00 & 372.60 & \textbf{338.69}\\
25 & 348.20 & 359.80 & 378.60 & 377.40 & 432.97 & 353.80 & 352.20 & \textbf{330.18}\\
30 & 346.20 & 356.20 & 362.20 & 375.80 & 431.46 & 349.80 & 351.40 & \textbf{325.74}\\
35 & 343.00 & 354.20 & 354.80 & 374.20 & 422.72 & 347.00 & 346.20 & \textbf{321.22}\\
\hline
\end{tabular}
\label{tab:14}
\end{table}


\begin{table}[h]
\scriptsize
\centering
\caption{Average best cost (in cm) per iteration count for Map 3}
\begin{tabular}{c c c c c c c c c}
\hline
Iteration & GWO & GSO & PSO & BBO & IBA & WOA & SCA &\textbf{SSA} \\
\hline
1 & 312.35 & 315.60 & 316.40 & 616.88 & 745.22 & 359.20 & 401.20 & \textbf{409.25}\\
5 & 272.60 & 274.00 & 304.80 & 356.00 & 449.28 & 282.00 & 314.00 & \textbf{341.50}\\
10 & 262.80 & 269.20 & 265.20 & 342.45 & 367.28 & 265.60 & 296.80 & \textbf{273.46}\\
15 & 257.60 & 264.00 & 263.60 & 283.20 & 358.20 & 258.20 & 261.20 & \textbf{244.60}\\
20 & 251.60 & 256.40 & 258.20 & 276.40 & 319.23 & 252.40 & 254.00 & \textbf{241.86}\\
25 & 250.80 & 246.80 & 250.00 & 276.00 & 271.60 & 252.20 & 251.60 & \textbf{239.90}\\
30 & 245.40 & 246.00 & 248.80 & 274.80 & 268.00 & 251.20 & 247.20 & \textbf{239.05}\\
35 & 245.00 & 245.80 & 248.60 & 272.80 & 266.40 & 251.20 & 247.00 & \textbf{237.65}\\
\hline
\end{tabular}
\label{tab:15}
\end{table}


\begin{table}[h]
\scriptsize
\centering
\caption{Average best cost (in cm) per iteration count for Map 4}
\begin{tabular}{c c c c c c c c c}
\hline
Iteration & GWO & GSO & PSO & BBO & IBA & WOA & SCA & \textbf{SSA} \\
\hline
1 & 476.80 & 461.20 & 496.05 & 525.80 & 518.60 & 530.0 & 481.60 & \textbf{471.60}\\
5 & 356.65 & 340.75 & 350.80 & 365.75 & 370.80 & 402.65 & 360.50 & \textbf{342.95}\\
10 & 295.40 & 310.90 & 308.50 & 310.95 & 340.95 & 338.70 & 334.85 & \textbf{289.45}\\
15 & 277.25 & 283.20 & 282.80 & 290.20 & 310.10 & 308.60 & 308.20 & \textbf{273.40}\\
20 & 276.85 & 280.45 & 276.40 & 279.50 & 296.40 & 299.65 & 293.80 & \textbf{271.25}\\
25 & 275.10 & 277.30 & 273.70 & 268.15 & 294.10 & 290.20 & 289.40 & \textbf{269.85}\\
30 & 274.85 & 275.65 & 271.65 & 258.90 & 292.45 & 285.60 & 286.00 & \textbf{269.10}\\
35 & 274.40 & 274.00 & 270.00 & 257.40 & 291.60 & 282.25 & 284.45 & \textbf{268.55}\\
\hline
\end{tabular}
\label{tab:16}
\end{table}



\begin{figure}[h]
\centering
\subfigure[Map 1]{\includegraphics[width=75mm,scale=0.50]{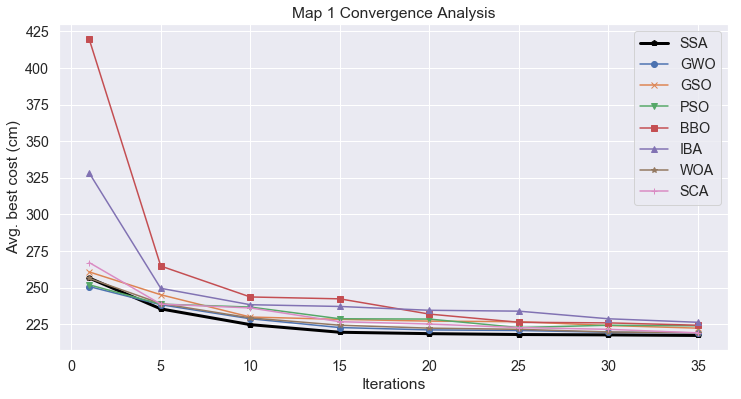}}
\hspace{1.0 mm}
\subfigure[Map 2]{\includegraphics[width=75mm,scale=0.50]{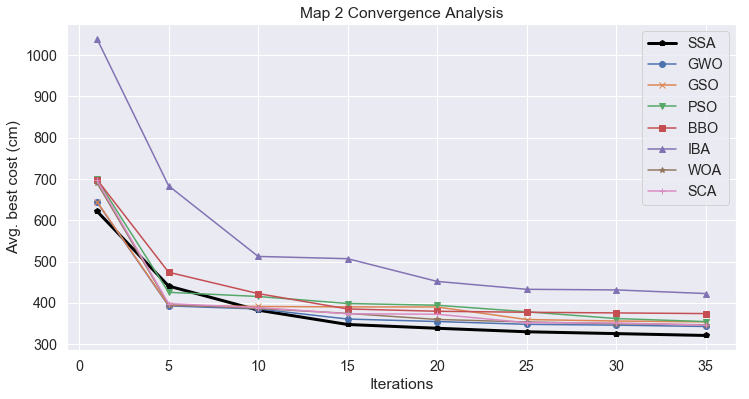}} \\
\subfigure[Map 3]{\includegraphics[width=75mm,scale=0.50]{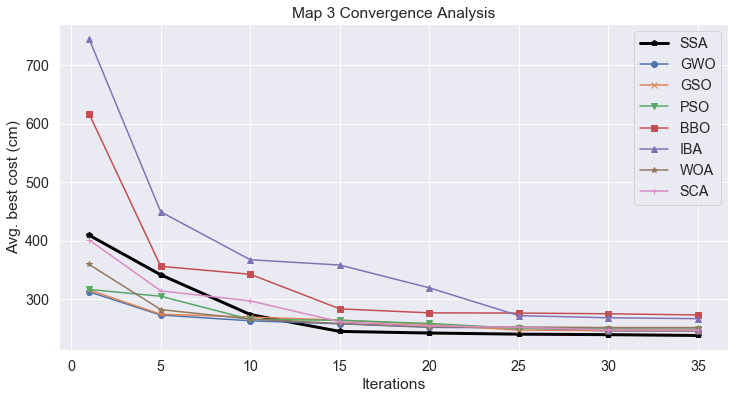}}
\hspace{1.0 mm}
\subfigure[Map 4]{\includegraphics[width=75mm,scale=0.50]{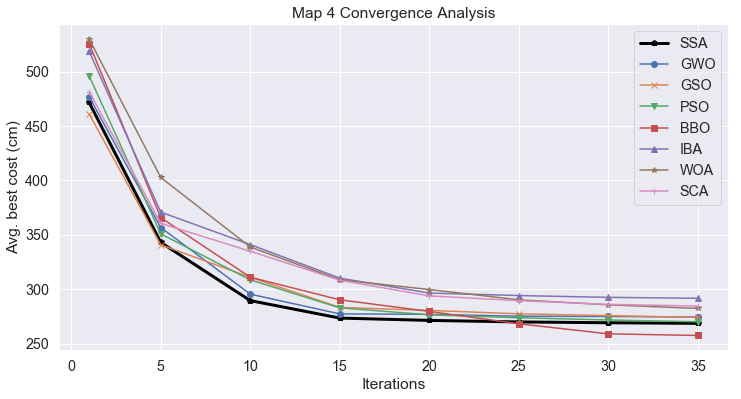}}
\caption{Convergence Curve of Different Maps}
\label{fig:9}
\end{figure}

As seen in Figure \ref{fig:9}, the execution time of algorithms are comparable to recently reported data \cite{ref-p}, when a simple environment like Map 1 and 2 is considered. For complex environment like Map 4, the SSA achieves low time complexity and more influencing results. Thus, SSA becomes suitable for real-life scenarios like route planning for multiple UAVs in a complex real-life environment. 

The average percentage improvement in cost and time of SSA algorithm with respect to GWO algorithm on Maps 1 to 4 as given in Tables 7 to 10, are calculated by equations 15 and 16.

\begin{equation}
\scriptsize
\frac{1}{4} \times \sum_{i = 1}^{4} \Bigg( \frac{1}{2} \times \sum_{j = 1}^{2} \bigg( \frac{1}{5} \times \frac{\sum\limits_{k = 1}^{5} Cost\; of\; GWO - \sum\limits_{k = 1}^{5} Cost\; of\; SSA}{\sum\limits_{k = 1}^{5} Cost\; of\; GWO} \times 100 \bigg) \Bigg)
\end{equation}

where $i$ = 1 to 4 represents Map 1, 2, 3 or 4, $j$ corresponds to results with respect to two different number of iterations for each map. The innermost summation in equation 15 calculates the effect of costs of all the five UAVs.


For Map 1, the improvement of SSA as compared to GWO is 1.04\%. Similarly for Maps 2, 3 and 4, the improvements are 0.81\%, 0.94\% and 2.21\%. Therefore using formula 15 we get the percentage average improvement in cost of SSA as 1.25\%.

\begin{equation}
\scriptsize
\frac{1}{4} \times \sum_{i = 1}^{4} \bigg( \frac{1}{2} \times \sum_{j = 1}^{2} \Big( \frac{Time\; of\; GWO - Time\; of\; SSA}{Time\; of\; GWO} \times 100 \Big) \bigg)
\end{equation}

The percentage improvement in time of SSA is computed to be is 9.40\%, 6.90\%, 2.27\% and 5.57\% for Maps 1, 2, 3 and 4 respectively. The percentage average improvement in time of SSA when compared to performance of GWO is 6.035\%.

\subsection{Comparison with other meta-heuristic algorithms}
Since there is no prior information regarding the best meta-heuristic, that is why various meta-heuristic algorithms are utilized for solving multiple UAVs route planning problem. Different algorithms like IBA, BBO, GSO, PSO, WOA, SCA, GWO, and SSA were tried, and it was observed that SSA has the best performance among them. SSA exhibits simplicity since it requires significantly fewer parameters, which are $c_1$, $c_2$ and $c_3$. This algorithm is also highly flexible since it is applicable to various types of problems and even its architecture need not to be changed. It has a gradient-free mechanism; thus, the need to calculate derivatives of search spaces is avoided. It optimizes the problem in stochastic fashion and prevents local optima stagnation which makes it suitable for solving multi-modal optimization problems. 

As stated in No Free Lunch (NFL) \cite{ref-v}, there is no particular meta-heuristic which is best for handling all optimization tasks. So a situation can arise, where a particular meta-heuristic might outperform all other algorithms significantly, but the
same meta-heuristic can severely give a bad performance for some other set of issues. 

\section{Conclusion and future scope}
\label{sec:6}

This paper proposes the use of SSA for solving multiple UAV route planning problem in a 3D environment. After investigating the performance of SSA in various experimental scenarios, it is concluded that SSA takes the least time in all the cases when compared with deterministic and other meta-heuristic algorithms and finds an optimal route in both 2D and 3D environment. In some of the cases, the route obtained by SSA has slightly more cost than the recently reported data \cite{ref-t}. However, after combining the overall effect of time and cost trade-off, it is realized that the time required in SSA is significantly less compared to other algorithms and thus accounts for the slightly higher cost and outperforms all other algorithms. So it becomes evident that SSA has the most superior performance than all other algorithms for fast, real-time, and optimal route re-planning. The simplicity, flexibility and gradient-free mechanism of the SSA make it immune to local optima stagnation and thereby improving the speed of convergence and making it suitable for the route planning and various other optimizations problems in real-life.

Future work may be focussed on extending this work by constructing an environment which better mimics a real-world scenario by introducing dynamic obstacles which, along with priority assignment associated with the goals and some hardware related constraints like minimum turning radius or maximum pitch angles should also be considered. A hybrid or a modified algorithm based on SSA for route planning and other real-world problem can be proposed to enhance the performance even further.


\begin{thebibliography}{}

\bibitem{ref-a} Mac T. T., et al.: `Heuristic approaches in robot path planning: A survey', \textit{Robotics and Autonomous Systems}, 2016, \textbf{86}, pp.~13--28

\bibitem{ref-b} Cai C., Silvia F.: `Information-driven sensor path planning by approximate cell decomposition', \textit{IEEE Transactions on Systems, Man, and Cybernetics, Part-B (Cybernetics)}, 2009, \textbf{39(3)}, pp.~672--689

\bibitem{ref-c} Barraquand J., Latombe J. C.: `Robot motion planning: A distributed representation approach', \textit{The International Journal of Robotics Research}, 1991, \textbf{10(6)}, pp.~628--649

\bibitem{ref-d} LaValle S. M.: `Rapidly-exploring random trees: A new tool for path planning', 1998

\bibitem{ref-e} Asadi S., et al.: `A novel global optimal path planning and trajectory method based on adaptive dijkstra-immune approach for mobile robot.' \textit{2011 IEEE/ASME Int. Conf. on Advanced Intelligent Mechatronics (AIM)}, 2011

\bibitem{ref-f} Kala R., Shukla A., Tiwari R.: `Fusion of probabilistic A* algorithm and fuzzy inference system for robotic path planning.' \textit{Artificial Intelligence Review}, 2010, \textbf{33(4)}, pp.~307--327

\bibitem{ref-g} Wu J.P., Peng Z. H., Chen J.: `3D multi-constraint route planning for UAV low-altitude penetration based on multi-agent genetic algorithm', \textit{IFAC Proceedings Volumes}, 2011, \textbf{44(1)}, pp.~11821--11826

\bibitem{ref-h} Zhang B., Duan H.: `Predator-prey pigeon-inspired optimization for UAV three-dimensional path planning', \textit{Int. Conf. in Swarm Intelligence.} Springer, Cham, 2014.

\bibitem{ref-i} Mirjalili S., Lewis A.: `The whale optimization algorithm', \textit{Advances in engineering software}, 2016, \textbf{95}, pp.~51--67

\bibitem{ref-j} Zhu W., Duan H.: `Chaotic predator–prey biogeography-based optimization approach for UCAV path planning', \textit{Aerospace science and technology}, 2014, \textbf{32(1)}, pp.~153--161

\bibitem{ref-k} Li S., Sun X., Xu Y.: `Particle swarm optimization for route planning of unmanned aerial vehicles', \textit{IEEE Int. Conf. on information acquisition.} IEEE, 2006

\bibitem{ref-l} Mohanty P. K., Parhi D. R.: `A new efficient optimal path planner for mobile robot based on Invasive Weed Optimization algorithm.' \textit{Frontiers of Mechanical Engineering}, 2014, \textbf{9(4)}, pp.~317--330

\bibitem{ref-m} Pandey P., Shukla A.,Tiwari R.: `Three-dimensional path planning for unmanned aerial vehicles using glowworm swarm optimization algorithm', \textit{International Journal of System Assurance Engineering and Management}, 2018, \textbf{9(4)}, pp.~836--852

\bibitem{ref-n} Hossain M. A., Ferdous I.: `Autonomous robot path planning in dynamic environment using a new optimization technique inspired by bacterial foraging technique', \textit{Robotics and Autonomous Systems}, 2015, \textbf{64}, pp.~137--141

\bibitem{ref-o} Dolicanin E., et al.: `Unmanned combat aerial vehicle path planning by brain storm optimization algorithm', \textit{Studies in Informatics and Control}, 2018, \textbf{27(1)}, pp.~15--24

\bibitem{ref-p} Zhang S., et al.: `Grey wolf optimizer for unmanned combat aerial vehicle path planning', \textit{Advances in Engineering Software}, 2016, \textbf{99}, pp.~121--136

\bibitem{ref-q} Phung M. D., et al.: `Enhanced discrete particle swarm optimization path planning for UAV vision-based surface inspection', \textit{Automation in Construction}, 2017, \textbf{81}, pp.~25--33

\bibitem{ref-r} Zhou Y., et al.: `A novel path planning algorithm based on plant growth mechanism', \textit{Soft Computing}, 2017, \textbf{21(2)}, pp.~435--445

\bibitem{ref-s} Mirjalili S., et al.: `Salp Swarm Algorithm: A bio-inspired optimizer for engineering design problems', \textit{Advances in Engineering Software}, 2017, \textbf{114} pp.~163--191

\bibitem{ref-t} Dewangan R. K., Shukla A., Godfrey W. W.: `Three dimensional path planning using Grey wolf optimizer for UAVs', \textit{Applied Intelligence}, 2019, \textbf{49(6)}, pp.~2201--2217

\bibitem{ref-u} Duan H., Ma G., Luo D. L.: `Optimal formation reconfiguration control of multiple UCAVs using improved particle swarm optimization', \textit{Journal of Bionic Engineering}, 2008, \textbf{5(4)} pp.~340--347

\bibitem{ref-v} Wolpert D. H., William G. M.: `No free lunch theorems for optimization', \textit{IEEE transactions on evolutionary computation}, 1997, \textbf{1(1)}, pp.~67--82

\bibitem{ref-w}Raja P., Pugazhenthi S.: `Optimal path planning of mobile robots: A review', \textit{International journal of physical sciences}, 2012, \textbf{7(9)}, pp.~1314--1320

\bibitem{ref-x}Tiwari R., Jain G.,Shukla A.: `MVO-Based Path Planning Scheme with Coordination of UAVs in 3-D Environment', \textit{Journal of Computational Science}, 2019

\end{thebibliography}
\end{document}